\documentclass[11pt]{article}
\usepackage{amssymb}
\usepackage{times}
\usepackage{amsmath}
\usepackage{epsfig}
\usepackage{naacl2001}

\newtheorem{theorem}{Theorem}

\newtheorem{claim}[theorem]{Claim}

\newenvironment{proof}[1][Proof]{\textbf{#1.} }{\ \rule{0.5em}{0.5em}}

\newcommand{\ignore}[1]{}

\newcommand{\ra}{\rightarrow}

\def\pp{\par\noindent}

\title{
A Sequential Model for Multi-Class Classification\thanks{This
research is supported by NSF grants IIS-9801638, IIS-0085836 and
SBR-987345.}}
\author{Yair Even-Zohar  \hspace{0.5in}  Dan Roth \\
     Department of Computer Science \\
         University of Illinois at Urbana-Champaign \\
         {\tt \{evenzoha,danr\}@uiuc.edu} }

\begin{document}
\maketitle

\begin{abstract}
Many classification problems require decisions among a
large number of competing classes.  These tasks, however, are not
handled well by general purpose learning methods and are usually
addressed in an ad-hoc fashion.
We suggest a general approach -- a sequential learning model that
utilizes classifiers to sequentially restrict the number of
competing classes while maintaining, with high probability, the
presence of the true outcome in the candidates set.
Some theoretical and computational properties of the model are
discussed and we argue that these are important in NLP-like
domains.
The advantages of the model are illustrated in an experiment in
part-of-speech tagging.
\end{abstract}

\section{Introduction} \label{sec:intro}


A large number of important natural language inferences can be
viewed as problems of resolving ambiguity, either semantic or
syntactic, based on properties of the surrounding context.
These, in turn, can all be viewed as classification problems in
which the goal is to select a class label from among a collection
of candidates.
Examples include part-of speech tagging, word-sense
disambiguation, accent restoration, word choice selection in
machine translation, context-sensitive spelling correction, word
selection in speech recognition and identifying discourse markers.

Machine learning methods have become the most popular technique
in a variety of classification problems of these sort, and have
shown significant success.
A partial list consists of Bayesian
classifiers~\cite{gale-word-sense}, decision
lists~\cite{Yarowsky94}, Bayesian hybrids~\cite{Golding95},
HMMs~\cite{Charniak93}, inductive logic methods~\cite{ZelleMo96},
memory-based methods~\cite{zdv-97}, linear
classifiers~\cite{Roth98,Roth99} and transformation-based
learning~\cite{Brill95}.

In many of these classification problems a significant source of
difficulty is the fact that the number of candidates is very large --
all words in words selection problems, all possible tags in tagging
problems etc.
Since general purpose learning algorithms do not handle these
multi-class classification problems well (see below), most of the
studies do not address the whole problem; rather, a small set of
candidates (typically two) is first selected, and the classifier
is trained to choose among these. While this approach is
important in that it allows the research community to develop
better learning methods and evaluate them in a range of
applications, it is important to realize that an important stage
is missing.
This could be significant when the classification methods are to
be embedded as part of a higher level NLP tasks
such as machine translation or information extraction, where the
small set of candidates the classifier can handle may not be
fixed and could be hard to determine.


In this work we develop a general approach to the study of
multi-class classifiers.  We suggest a sequential learning model
that utilizes (almost) general purpose classifiers to
sequentially restrict the number of competing classes while
maintaining, with high probability, the presence of the true
outcome in the candidate set.

In our paradigm the sought after classifier has to choose a
single class label (or a small set of labels) from among a large
set of labels. It works by sequentially applying simpler
classifiers, each of which outputs a probability distribution over
the candidate labels.
These distributions are {\em multiplied} and thresholded,
resulting in that each classifier in the sequence needs to deal
with a (significantly) smaller number of the candidate labels
than the previous classifier.
The classifiers in the sequence are selected to be simple in the
sense that they typically work only on part of the feature space
where the decomposition of feature space is done so as to achieve
statistical independence.
%
Simple classifier are used since they are more likely to be
accurate; they are chosen so that, with high probability
(w.h.p.), they have one sided error, and therefore the presence
of the true label in the candidate set is maintained. The order of
the sequence is determined so as to maximize the rate of
decreasing the size of the candidate labels set.

Beyond increased accuracy on multi-class classification problems
, our scheme improves the computation time of these problems
several orders of magnitude, relative to other standard schemes.

In this work we describe the approach, discuss an experiment done
in the context of part-of-speech (pos) tagging, and provide some
theoretical justifications to the approach.
Sec.~\ref{sec:background} provides some background on approaches
to multi-class classification in machine learning and in NLP. In
Sec.~\ref{sec:sm} we describe the sequential model proposed here
and in Sec.~\ref{sec:exp} we describe an experiment the exhibits
some of its advantages. Some theoretical justifications are
outlined in Sec.~\ref{sec:theory}.


\section{Multi-Class Classification} \label{sec:background}
Several works within the machine learning community have
attempted to develop general approaches to multi-class
classification.  One of the most promising approaches is that of
error correcting output codes~\cite{DiettrichBa95a}; however,
this approach has not been able to handle well a large number of
classes (over 10 or 15, say) and its use for most large scale NLP
applications is therefore questionable.
Statistician have studied several schemes such as learning a
single classifier for each of the class labels ({\it one vs.
all}) or learning a discriminator for each pair of class labels,
and discussed their relative merits\cite{HastieTi98}. Although it
has been argued that the latter should provide better results than
others, experimental results have been mixed~\cite{AllweinScSi00}
and in some cases, more involved schemes, e.g., learning a
classifier for each set of {\em three} class labels (and deciding
on the prediction in a tournament like fashion) were shown to
perform better~\cite{TeowLo00}.
Moreover, none of these methods seem to be computationally
plausible for large scale problems, since the number of
classifiers one needs to train is, at least, quadratic in the
number of class labels.


Within NLP, several learning works have already addressed the
problem of multi-class classification.  In~\cite{KudohMa00} the
methods of ``all pairs'' was used to learn phrase annotations for
shallow parsing. More than $200$ different classifiers where used
in this task, making it infeasible as a general solution.
%
%
All other cases we know of, have taken into account some properties of
the domain and, in fact, several of the works can be viewed as
instantiations of the sequential model we formalize here, albeit done
in an ad-hoc fashion.

In speech recognition, a sequential model is used to process speech
signal. Abstracting away some details, the first classifier used is a
speech signal analyzer; it assigns a positive probability only to some
of the words (using Levenshtein distance~\cite{Levenshtein66} or
somewhat more sophisticated techniques~\cite{Levinson90}).
These words are then assigned probabilities using a different
contextual classifier e.g., a language model,
and then, (as done in most current speech recognizers) an
additional sentence level classifier uses the outcome of the word
classifiers in a word lattice to choose the most likely sentence.
%

Several word prediction tasks make decisions in a sequential way
as well.  In spell correction {\em confusion sets} are created
using a classifier that takes as input the word transcription and
outputs a positive probability for potential words.
In conventional spellers, the output of this classifier is then
given to the user who selects the intended word.
In context sensitive spelling correction~\cite{GoldingRo99,ManguBr97}
an additional classifier is then utilized to predict among words that
are supported by the first classifier, using contextual and lexical
information of the surrounding words. In all studies done so far,
however, the first classifier -- the confusion sets -- were
constructed manually by the researchers.

Other word predictions tasks have also constructed manually the
list of confusion sets~\cite{LeePe99b,DaganLePe99,Lee99} and
justifications where given as to why this is a reasonable way to
construct it.  \cite{EvenZoharRo00} present a similar task in
which the confusion sets generation was automated. Their study
also quantified experimentally the advantage in using early
classifiers to restrict the size of the confusion set.

Many other NLP tasks, such as pos tagging, name entity
recognition and shallow parsing require multi-class classifiers.
In several of these cases the number of classes could be very
large (e.g., pos tagging in some languages, pos tagging when a
finer proper noun tag is used). The sequential model suggested
here is a natural solution.


\section{The Sequential Model} \label{sec:sm}
We study the problem of learning a multi-class classifier,
$f:X\rightarrow C$ where $X\subseteq \{0,1\}^n$,
$C=\{c_1,...,c_m\}$
and $m$ is typically large, on the order of $10^2-10^5$.
We address this problem using the Sequential Model (SM) in which
simpler classifiers  are sequentially used to filter subsets of $C$ out
of consideration.
%
%

The sequential model is formally defined as a $5$-tuple:
$$SM = \{~\{X^i\},  ~C, ~O, ~\{f_i\}, ~\{\epsilon_i\}~\},$$
where

\begin{itemize}
\item
$X=\cup_{i=1}^{N}X^i$ is a decomposition of the domain (not
necessarily disjoint; it could be that $\forall i, X^i=X$).
\item
$C$ is the set of class labels.
\item
$O=\{o_1,o_2,...,o_{N}\}$ determines the order in which the
classifiers are learned and evaluated. For convenience we denote
$f_1=f_{o_1}, f_2=f_{o_2},\ldots$
\item
$\{f_i\}_1^N$ is the set of classifiers used by the model,
$f_i:(X^i, 2^{|C|})\rightarrow [0,1]^{|C|}$.

\item
$\{\epsilon_i\}_1^N$ is a set of constant thresholds.
\end{itemize}
Given $x \in X^i$ and a set $C_{i-1}$ of class labels, the $i$th
classifier outputs a probability distribution\footnote{The output of
many classifiers can be viewed, after appropriate normalization, as
a confidence measure that can be used as our $P_i$.}
$P_i =(p_i(c_{1}|x),...,p_i(c_{m}|x))$ over labels in $C$ (where
$p_i(c|x)$ is the probability assigned to class $c$ by $f_i$), and
$P_i$ satisfies that if $c \notin C_{i-1}$ then $p_i(c|x)=0$.

The set of remaining candidates after the $i$th classification
stage is determined by $P_i$ and $\epsilon_i$:
$$C_i = \{ c \in C | p_i(c|x)
> \epsilon_i \}.$$

The sequential process can be viewed as a multiplication of
distributions. \cite{Hinton00} argues that a {\em product} of
distributions (or, ``experts'', PoE) is an efficient way to make
decisions in cases where several different constrains play a
role, and is advantageous over additive models.
In fact, due to the thresholding step, our model can be viewed as
a selective PoE.
The thresholding ensures that the SM has the following
monotonicity property:
$$
\hspace{-0.04in}%
 \{c\in C| ~p_i(c|x)>\epsilon_i\}\subseteq \hspace{-0.02in} \{c\in
C| ~p_{i-1}(c|x)>\epsilon_{i-1}\}
$$
%
that is, as we evaluate the classifiers sequentially, smaller or
equal (size) confusion sets are considered.
A desirable design goal for the SM is that, w.h.p., the
classifiers have one sided error (even at the price of rejecting
fewer classes). That is, if $c_t$ is the true target\footnote{We
use the terms class and target interchangeably.}, then we would
like to have that $p_i(c_t|x)>\epsilon_i$.
The rest of this paper presents a concrete instantiation of the
SM, and then provides a theoretical analysis of some of its
properties (Sec.~\ref{sec:theory}). This work does not address the
question of acquiring SM i.e., learning $\{\epsilon_i\}, O$.

%
%
%



%
%
\section{Example: POS Tagging} \label{sec:exp}
This section describes a two part experiment of pos tagging in
which we compare, under identical conditions, two classification
models: A SM and a single classifier. Both are provided with the
same input features and the only difference between them is the
model structure.

In the first part, the comparison is done in the context of
assigning pos tags to unknown words -- those words which were not
presented during training and therefore the learner has no
baseline knowledge about possible POS they may take. This
experiment emphasizes the advantage of using the SM during
evaluation in terms of accuracy.
The second part is done in the context of pos tagging of known
words. It compares processing time as well as accuracy of
assigning pos tags to known words (that is, the classifier
utilizes knowledge about possible POS tags the target word may
take). This part exhibits a large reduction in training time using
the SM over the more common {\it one-vs-all} method while the
accuracy of the two methods is almost identical.
%




Two types of features -- {\tt lexical} features and {\tt
contextual} features may be used when learning how to tag words
for pos. Contextual features capture the information in the
surrounding context and the word lemma while the lexical features
capture the morphology of the unknown word.\footnote{Lexical
features are used only when tagging unknown words.}
Several issues make the pos tagging problem a natural problem to
study within the SM.
(i) A relatively large number of classes (about 50).
(ii) A natural decomposition of the feature space to contextual
and lexical features.
(iii) Lexical knowledge (for unknown words) and the word lemma
(for known words) provide, w.h.p, one sided error
\cite{Mikheev97}.

%
\subsection{The Tagger Classifiers} \label{sec:tag}

The domain in our experiment is defined using the following set of
features, all of which are computed relative to the target word
$w_i$.
\subsubsection*{Contextual Features (as in \cite{Brill95,RothZe98}):}
Let $t_{i-1}, (t_{i+1})$ be the tags of the word preceding,
(following) the target word, respectively.

 1. $t_{i-1}$.

2. $t_{i+1}$.

3. $t_{i-2}$.

4. $t_{i+2}$.

5. $t_{i-1}\&t_{i+1}$.

6. $t_{i-2}\&t_{i-1}$.

7. $t_{i+1}\&t_{i+2}$.

8. Baseline tag for word $w_i$. In case $w_i$ is an unknown word,
the baseline is {\it proper singular noun ``NNP''} for capitalized
words and {\it common singular noun ``NN''} otherwise. (This
feature is introduced only in some of the experiments.)

9.The target word $w_i$.

\subsubsection*{Lexical Features:}

Let $\alpha, \beta, \gamma$ be any three characters observed in
the examples.

10. Target word is capitalized.

11. $w_i$ ends with $\alpha$ and length($w_i)
> 3$.

12. $w_i$  ends with $\beta \alpha$ and length($w_i) > 4$.

13. $w_i$  ends with $\gamma \beta \alpha$ and length($w_i)
> 5$.

\begin{figure}[h]
\begin{center}
\def\baselinestretch{1}
\hspace {-0.1in}
\epsfig{file=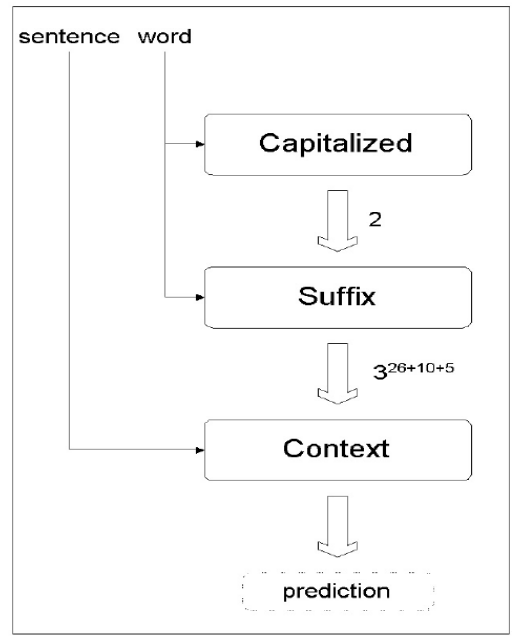,height=3.3695in,width=2.83481in}
\caption{\textbf{POS Tagging  of Unknown Word using  Contextual
and Lexical features in a Sequential Model. The input for
capitalized classifier has 2 values and therefore 2 ways to create
confusion sets. There are at most $3^{(26+10+5)}$ different
inputs for the suffix classifier (26 character + 10 digits + 5
other symbols), therefore suffix may emit up to $3^{(26+10+5)}$
confusion sets.} \label{fig:unknown}}
\end{center}\def\baselinestretch{1.3}  
\end{figure}

In the following experiment, the SM used for unknown words makes
use of three different classifiers $f_1, f_2$ and $f_3$ or
$f_3'$, defined as follows:

\begin{description}
\item[$f_1=$:]
a classifier based on the lexical feature $\#10$.
\item[$f_2=$:] a classifier based on lexical features $\#11-13$
\item[$f_3=$:]
a classifier based on contextual features $\#1-9$.
\item[$f_3'=$:]
a classifier based on all the features, $\#1-13$.
\end{description}
The SM is compared with a single classifier -- either $f_3$ or
$f_3'$. Notice that $f_{3}'$ is a single classifier that uses the
same information as used by the SM.
%
Fig~\ref{fig:unknown} illustrates the SM that was used in the
experiments.

All the classifiers in the sequential model, as well as the
single classifier, use the SNoW learning architecture~\cite{Roth98}
with the Winnow update rule.
SNoW (Sparse Network of Winnows) is a multi-class classifier that
is specifically tailored for learning in domains in which the
potential number of features taking part in decisions is very
large, but in which decisions actually depend on a small number
of those features. SNoW works by learning a sparse network of
linear functions over a pre-defined or incrementally learned
feature space.
SNoW has already been used successfully on several tasks in
natural language
processing~\cite{Roth98,RothZe98,GoldingRo99,PR01nips}.

Specifically, for each class label SNoW learns a function $f_c:X
\ra [0,1]$ that maps a feature based representation $x$ of the
input instance to a number $a_c(x) \in [0,1]$ which can be
interpreted as the probability of $c$ being the class label
corresponding to $x$. At prediction time, given $x \in X$, SNoW
outputs
\begin{equation} \label{eq:snow}
SNoW(x) = max_{c} \{a_c(x)\}.
\end{equation}

All functions -- in our case, $50$ target nodes are used, one for each
pos tag -- reside over the same feature space, but can be thought of
as autonomous functions (networks). That is, a given example is
treated autonomously by each target subnetwork; an example labeled $t$
is considered as a positive example for the function learned for $t$
and as a negative example for the rest of the functions (target
nodes). The network is {\em sparse} in that a target node need not be
connected to all nodes in the input layer. For example, it is not
connected to input nodes (features) that were never active with it in
the same sentence.

Although SNoW is used with $50$ different targets, the SM utilizes by
determining the confusion set dynamically. That is, in evaluation
(prediction), the {\em maximum} in Eq.~\ref{eq:snow} is taken only
over the currently applicable confusion set. Moreover, in training, a
given example is used to train only target networks that are in the
currently applicable confusion set. That is, an example that is
positive for target $t$, is viewed as positive for this target (if it
is in the confusion set), and as negative for the other targets in the
confusion set. All other targets do not see this example.

The case of POS tagging of known words is handled in a similar way.
In this case, all possible tags are known. In training, we record, for
each word $w_i$, all pos tags with which it was tagged in the training
corpus. During evaluation, whenever word $w_i$ occurs, it is tagged
with one of these pos tags. That is, in evaluation, the confusion set
consists only of those tags observed with the target word in training,
and the {\em maximum} in Eq.~\ref{eq:snow} is taken only over
these. This is always the case when using $f_3$ (or $f_{3}'$), both in
the SM and as a single classifier.
In training, though, for the sake of this experiment, we treat
$f_3$ ($f_{3}'$) differently depending on whether it is trained
for the SM or as a single classifier.
When trained as a single classifier (e.g., \cite{RothZe98}),
$f_3$ uses each $t$-tagged example as a positive example for $t$
and a negative example for all other tags. On the other hand, the
SM classifier is trained on a $t$-tagged example of word $w$, by
using it as a positive example for $t$ and a negative example
only for the effective confusion set. That is, those pos tags which have
been observed as tags of $w$ in the training corpus.


\subsection{Experimental Results}

The data for the experiments was extracted from the Penn Treebank
WSJ  and Brown corpora. The training corpus consists of
$2,400,000$ words. The test corpus consists of $280,000$ words of
which $5,412$ are unknown words (that is, they do not occur in
the training corpus. (Numbers (the pos ``CD''), are not included
among the unknown words).


\subsubsection*{POS Tagging of Unknown Words}
\begin{table}[h] \centering
\def\baselinestretch{1}  
\begin{tabular}{|c|c|c|}
\hline   $f_3$ & $f_3$ + baseline  & baseline\\
\hline   $8.6$ & $61.8$               & $60.8$ \\
\hline
\end{tabular}
\caption{{\bf POS tagging of unknown words using contextual
features (accuracy in percent)}. $f_3$ is a classifier that uses
only contextual features, $f_3$ + baseline is the same
classifier with the addition of the baseline feature (``NNP'' or
``NN''). }

\label{tab:base-results}
\end{table}

Table~\ref{tab:base-results} summarizes the results of the
experiments with a single classifier that uses only contextual
features. Notice that adding the baseline POS significantly
improves the results but not much is gained over the baseline.
The reason is that the baseline feature is almost perfect
($94.4\%$) in the training data.
For that reason, in the next experiments we do not use the
baseline at all, since it could hide the phenomenon addressed.
(In practice, one might want to use a more sophisticated
baseline, as in \cite{Dermatas95}.)

\begin{table}[h] \centering
\def\baselinestretch{1}  
\begin{tabular}{|c|c|c|c|}
\hline   $f_3$ & $f_3'$ & \textbf{SM($f_1,f_2,f_3$)} & \textbf{SM($f_1,f_2,f_3'$)}\\
\hline   $8.6$         & $56.1$              & $65.7$                    & $73.0$ \\
\hline
\end{tabular}
\caption{{\bf POS tagging of unknown words using contextual and
lexical Features (accuracy in percent)}. $f_3$ is based only on
contextual features, $f_3'$ is based on contextual and lexical
features. SM($f_i,f_j$) denotes that $f_j$ follows $f_i$ in the
sequential model.} \label{tab:results}
\end{table}

Table~\ref{tab:results} summarizes the results of the main
experiment in this part.
It exhibits the advantage of using the SM (columns 3,4) over a
single classifier that makes use of the same features set (column
2). In both cases, all features are used. In $f_3'$, a classifier
is trained on input that consists of all these features and
chooses a label from among all class labels.
In $SM(f_1,f_2,f_3)$ the same features are used as input, but
different classifiers are used sequentially -- using only part of
the feature space and restricting the set of possible outcomes
available to the next classifier in the sequence -- $f_i$ chooses
only from among those left as candidates.

It is interesting to note that further improvement can be
achieved, as shown in the right most column.  Given that the last
stage in $SM(f_1,f_2,f_3')$ is identical to the single classifier
$f_{3}'$, this shows the contribution of the filtering done in
the first two stages using $f_1$ and $f_2$.  In addition, this
result shows that the input spaces of the classifiers need not be
disjoint.

\subsubsection*{POS Tagging of Known Words}
Essentially everyone who is learning a POS tagger for known words
makes use of a ``sequential model'' assumption during evaluation
-- by restricting the set of candidates, as discussed in
Sec~\ref{sec:tag}).
The focus of this experiment is thus to investigate the advantage
of the SM during training. In this case, a single ({\it
one-vs-all}) classifier trains each tag against {\em all} other
tags, while a SM classifier trains it only against the effective
confusion set (Sec~\ref{sec:tag}).


Table~\ref{tab:known-results} compares the performance of the
$f_3$ classifier trained using in a {\it one-vs-all} method to the
same classifier trained the SM way. The results are only for
known words and the results of Brill's tagger~\cite{Brill95} are
presented for comparison.

\begin{table}[h] \centering
\def\baselinestretch{1}  
\begin{tabular}{|c|c|c|}
\hline   \hspace{ 0.05in} {\it one-vs-all}\hspace{ 0.05in}  & \hspace{ 0.05in}SM$_{train}$\hspace{ 0.05in} & \hspace{ 0.05in} Brill\hspace{ 0.05in}\\
\hline   $96.88$           & $96.86$ & $96.49$ \\
\hline
\end{tabular}
\caption{{\bf POS Tagging of known words using contextual features
(accuracy in percent)}. {\it one-vs-all} denotes training where
example $x$ serves as positive example to the true tag and as
negative example to all the other tags. SM$_{train}$ denotes
training where example $x$ serves as positive example to the true
tag and as a negative example only to a restricted set of tags in
based on a previous classifier -- here, a simple baseline
restriction.}
\label{tab:known-results}
\end{table}
While, in principle, (see Sec~\ref{sec:theory}) the SM should do
better (an never worse) than the one-vs-all classifier, we believe
that in this case SM does not have any performance advantages
since the classifiers work in a very high dimensional feature
space which allows the one-vs-all classifier to find a separating
hyperplane that separates the positive examples many different
kinds of negative examples (even irrelevant ones).

However, the key advantage of the SM in this case is the
significant decrease in computation time, both in training and
evaluation.
Table~\ref{tab:process-time} shows that in the pos tagging task,
training using the SM is {\bf 6} times faster than with a {\it
one-vs-all} method and {\bf 3000} faster than Brill's learner. In
addition, the evaluation time of our tagger was about twice
faster than that of Brill's tagger.
\begin{table}[h] \centering
\def\baselinestretch{1}  
\begin{tabular}{|c|c c|c|}
\hline            & \makebox[0.5in][r]{ {\it one-vs-all }} \vline &\hspace{-0.3in} SM$_{train}$ & Brill\\
\hline   {\bf Train}   & \makebox[0.5in]{$1877.3 $}   \vline &\hspace{-0.3in} $313.5$ & $ > 10^6$ \\
\hline   {\bf Test} & \makebox[1in][r]{ $ 2.3*10^{-3}$} &  & $4.3*10^{-3}$ \\
\hline
\end{tabular}
\caption{{\bf Processing time for POS tagging of known words using
contextual features (In CPU seconds)}. Train: training time over
$10^5$ sentences. Brill's learner was interrupted after 12 days of
training (default threshold was used). Test: average number of
seconds to evaluate a single sentence. All runs were done on the
same machine.}
\label{tab:process-time}
\end{table}


\section{The Sequential model: Theoretical Justification} \label{sec:theory}
In this section, we discuss some of the theoretical aspects of the
SM and explain some of its advantages. In particular, we discuss
the following issues:
\begin{enumerate}
\item
Domain Decomposition: When the input feature space can be
decomposed, we show that it is advantageous to do it and learn
several classifiers, each on a smaller domain.

\item
Range Decomposition: Reducing confusion set size is advantageous both in
training and testing the classifiers.
\begin{enumerate}
\item
Test: Smaller confusion set is shown to yield a smaller expected
error.
%
\item
Training: Under the assumptions that a small confusion set
(determined dynamically by previous classifiers in the sequence)
is used when a classifier is evaluated, it is shown that training
the classifiers this way is advantageous.

\end{enumerate}
\item
Expressivity:
SM can be viewed as a way to generate an expressive classifier by
building on a number of simpler ones. We argue that the SM way of
generating an expressive classifier has advantages over other
ways of doing it, such as decision tree. (Sec~\ref{sec:DT}).
\end{enumerate}

In addition, SM has several significant computational advantages
both in training and in test, since it only needs to consider a
subset of the set of candidate class labels. We will not discuss
these issues in detail here.

\subsection{Decomposing the Domain}\label{sec:prob}
Decomposing the domain is not an essential part of the SM; it is
possible that all the classifiers used actually use the same
domain. As we shown below, though, when a decomposition is
possible, it is advantageous to use it.

It is shown in Eq.~\ref{eq:ML0}-\ref{eq:ML5} that when it is
possible to decompose the domain to subsets that are conditionally
independent given the class label, the SM with classifiers
defined on these subsets is as accurate as the optimal single
classifier. (In fact, this is shown for a pure product of simpler
classifiers; the SM uses a selective product.)

%
%
%
%

In the following we assume that $X^1,\ldots,X^N$ provide a
decomposition of the domain $X$ (Sec.~\ref{sec:sm}) and that
$(x^1,\ldots,x^N) \in (X^1,\ldots,X^N)$.
By conditional independence we mean that
$$\forall i,j~~p(x^i,...,x^j|c) =
\prod_{k=i}^{j}p(x^k|c),$$
where $x^k$ is the input for the $k$th
classifier.
\begin{align}
&\underset{c\in C}{\arg \max }\;p(c|x)=\underset{c\in C}{\arg
\max }\;p(c|x^1,...,x^{N})
\label{eq:ML0}\\
& =\underset{c\in C}{\arg \max }\;\frac{p(x^1,...,x^{N}|c)\cdot
p(c)}{p(x^1,...,x^{N})}
\label{eq:ML1} \\
& =\underset{c\in C}{\arg \max }\;p(x^1,...,x^{N}|c)\cdot p(c)
\label{eq:ML2} \\
& =\underset{c\in C}{\arg \max }\;p(x^1|c)\cdots p(x^{N}|c)
\cdot p(c) \label{eq:ML3} \\
& =\underset{c\in C}{\arg \max }\;\frac{p(c|x^1)p(x^1)}{p(c)}
\cdots \frac{p(c|x^{N})p(x^{N})}{p(c)} \cdot p(c)
\label{eq:ML4}\\
& =\underset{c\in C}{\arg \max }\;p(c|x^1)\cdots p(c|x^{N}) \cdot
\frac {1}{{p(c)}^{N-1}} \label{eq:ML5}
\end{align}
$p(x^1,...,x^{N})$ in Eq.~\ref{eq:ML1} is identical $\forall c\in
C$ and therefore can be treated as a constant. Eq.~\ref{eq:ML3} is
derived by applying the independence assumption. Eq.~\ref{eq:ML4}
is derived by using the Bayes rule for each term $p(c|x^i)$
separately.

We note that although the conditional independence assumption is
a strong one, it is a reasonable assumption in many NLP
applications; in particular, when cross modality information is
used, this assumption typically holds for decomposition that is
done across modalities. For example, in POS tagging, lexical
information is often conditionally independent of contextual
information, given the true POS. (E.g., assume that word is a
gerund; then the context is independent of the ``ing'' word
ending.)

In addition, decomposing the domain has significant advantages
from the learning theory point of view~\cite{Roth99}. Learning
over domains of lower dimensionality implies better generalization
bounds or, equivalently, more accurate classifiers for a fixed
size training set.

\subsection{Decomposing the range} \label{sec:range}
The SM attempts to reduce the size of the candidates set. We
justify this by considering two cases: (i) Test: we will argue
that  prediction among a smaller set of classes has advantages
over predicting among a large set of classes;
(ii) Training: we will argue that it is advantageous to ignore
irrelevant examples.

\subsubsection{Decomposing the range during Test}
The following discussion formalizes the intuition that a smaller
confusion set in preferred. Let $f:X\rightarrow C$ be the true
target function and $p(c_j|x)$ the probability assigned by the
final classifier to class $c_j\in C$ given example $x\in X$.
Assuming that the prediction is done, naturally, by choosing the
most likely class label, we see that the expected error when
using a confusion set of size $k$ is:
\begin{align}
Error_k & = E_x[(\underset{1\leq j\leq k}{argmax}\;p(c_j|x)) \neq
f(x)] \notag\\
& = p((\underset{1\leq j\leq k}{argmax}\;p(c_j|x)) \neq f(x))
\end{align}
Now we have:
\begin{claim}\label{cl:test}
Let $K=\{ c_1,...,c_k\}, K'=\{c_1,...,c_{k+r}\}$ be two sets of
class labels and assume $f(x)\in K$ for example $x$. Then
$Error_k \leq Error_{k'}$.
\end{claim}

\begin{proof}
Denote:
$$pe(a,b,f)=p((\underset{a\leq j\leq b}{argmax}\;p(c_j|x)) \neq f(x))$$
Then,
\begin{align}
&Error_{K'}  = \notag\\
& = E_x[(\underset{1\leq j\leq k+r}{argmax}\;p(c_j|x))
\neq f(x)] \notag\\
& = pe(1,k+r,f) \notag\\
& = pe(1,k,f)\hspace{-0.05in} +( 1\hspace{-0.05in} - pe(1,k,f))
pe(k+1,k+r,f)\notag\\
& = Error_{K} + (1\hspace{-0.05in} -Error_{K})
pe(k+1,k+r,f) \notag\\
& \geq Error_{K} \notag
\end{align}
\end{proof}

Claim~\ref{cl:test} shows that reducing the size of the confusion
set can only help; this holds under the assumption that the true
class label is not eliminated from consideration by down stream
classifiers, that is, under the one-sided error assumption.
Moreover, it is easy to see that the proof of Claim~\ref{cl:test}
allows us to relax the one sided error assumption and assume
instead that the previous classifiers err with a probability
which is smaller than:
$$(1-Error_{K})\cdot pe(k+1,k+r,f(x)).$$

\subsubsection{Decomposing the range during training}

We will assume now, as suggested by the previous discussion, that
in the evaluation stage the smallest possible set of candidates
will be considered by each classifier. Based on this assumption,
Claim~\ref{cl:train} shows that training this way is advantageous.
That is, that utilizing the SM in training yields a better
classifier.

Let $\cal{A}$ be a {\it learning algorithm} that is trained to
minimize:
$$\int_{x\in X} L(y\cdot h(x))p(x)dx,$$
where $x$ is an example, $y\in \{-1,+1\}$ is the true class, $h$
is the hypothesis, $L$ is a loss function and $p(x)$ is the
probability of seeing example $x$ when $x\sim P$ (see
~\cite{AllweinScSi00}). (Notice that in this section we are using
general loss function $L$; we could use, in particular, binary
loss function used in Sec~\ref{sec:range}.) We phrase and prove
the next claim, w.l.o.g, the case of $2$ vs. $3$ class labels.
\begin{claim}\label{cl:train}
Let $C=\{c_1,c_2,c_3\}$ be the set of class labels, let $S_i$ be
the set of examples for class $i$.
Assume a sequential model in which class $c_1$ does not compete
with class $c_3$. That is, whenever $x\in S_1$ the SM filters out
$c_3$ such that the final classifier ($f_N$) considers only $c_1$
and $c_2$.
Then, the error of the hypothesis - produced by algorithm
$\cal{A}$ (for $f_N$) - when trained on examples in $\{S_1, S_2\}$
is no larger than the error produced by the hypothesis it
produces when trained on examples in $\{S_1, S_2, S_3\}$.
\end{claim}

\begin{proof}
Assume that the algorithm $\cal{A}$, when trained on a sample $S$,
produces a hypothesis that minimizes the empirical error over $S$.

Denote $x \sim P_{C}$ when $x$ is sampled according to a
distribution that supports only examples with label in $C$.
Let $S$ be a sample set of size $m$, according to $P_{1,2}$, and
$h'$ the hypothesis produced by $\cal{A}$. Then, for all $h \not
= h'$,

\begin{equation}\label{eq:dis-margin}
\frac{1}{m}\sum_{x\in S} L(yh'(x))\leq \frac{1}{m}\sum_{x\in S}
L(yh(x))
\end{equation}
In the limit, as $m\rightarrow \infty$
$$\underset{x\sim P_{1, 2}}\int \hspace{-0.1in} L(yh'(x))p(x)dx\leq
\underset{x\sim P_{1, 2}}\int \hspace{-0.1in} L(yh(x))p(x)dx.$$

In particular this holds if $h$ is a hypothesis produced by
$\cal{A}$ when trained on $S'$, that is sampled according to
$x\sim P_{1,2,3}$.
\end{proof}


\subsection {Expressivity}\label{sec:DT}
The SM is a decision process that is conceptually similar to a
decision tree processes~\cite{Safavian91,Mitchell97},
especially if one allows more general
classifiers in the decision tree nodes.
In this section we show that (i) the SM can express any DT.  (ii) the
SM is more compact than a decision tree even when the DT makes used of
more expressive internal nodes~\cite{MurthyKaSa94}.
%
%


The next theorem
shows that for a fixed set of functions (queries) over the input
features, any binary decision tree can be represented as a
SM. Extending the proof beyond binary decision trees is
straight-forward.

\begin{theorem}\label{th:DTinSM}
Let $T$ be a binary decision tree with $N$ internal nodes.  Then,
there exist a sequential model $S$ such that $S$ and $T$ have the same
size, and they produce the same predictions.
\end{theorem}
%

\pp {\bf Proof (Sketch):} Given a decision tree $T$ on $N$ nodes we show
how to construct a SM that produces equivalent predictions.

\begin{enumerate}
\item
Generate a confusion set $C$ the consists of $N$ classes, each
representing an internal node in $T$.

\item For each
internal node in $d\in T$, assign a classifier:
$f_i:X\times C\rightarrow [0,1]^{m-1+M}$.

\item Order the classifiers $f_1,...f_N$ such that a classifier that is
assigned to node $d$ is processed before any classifier that was
assigned to any of the children of $d$.

\item Define each classifier $f_i$ that was assigned to node $d\in T$
to have an influence on the outcome {\it iff} node $d\in T$ lies in
the path ($b_0,b_1,...,b_{k-1}$) from the root to the predicted class.

\item Show that using steps 1-4, the predicted target of $T$ and $S$
are identical.
\end{enumerate}
This completes that proof and shows that the resulting SM is of
equivalent size to the original decision tree.

We note that given a SM, it is also relatively easy (details omitted)
to construct a decision tree that produces the same decisions as the
final classifier of the SM.
However, the simple construction results in a decision tree that is
exponentially larger than the original SM.  Theorem~\ref{thrm:DT}
shows that this difference in expressivity is inherent.

\begin{theorem} \label{thrm:DT}
Let $N$ be the number of classifiers in a sequential model $S$ and the
number of internal nodes a in decision tree $T$. Let $m$ be the set of
classes in the output of $S$ and also the maximum degree of the
internal nodes in $T$.
Denote by $F(T), F(S)$ the number of functions representable by
$T, S$ respectively.
Then, when $m>>N$, $F(S)$ is exponentially larger than $F(T)$.
\end{theorem}

{\bf Proof (Sketch):} The proof follows by counting the number of
functions that can be represented using a decision tree with $N$
internal nodes\cite{Wilf94}, and the number of functions that can be
represented using a sequential model on $N$ intermediate classifier.
Given the exponential gap, it follows that one may need exponentially
large decision trees to represent an equivalent predictor to an $N$
size SM.

\section{Conclusion} \label{sec:conc}

A wide range and a large number of classification tasks will have to
be used in order to perform any high level natural language inference
such as speech recognition, machine translation or question answering.
Although in each instantiation the {\em real} conflict could be only to
choose among a small set of candidates,
the original set of candidates could be very large; deriving the small
set of candidates that are relevant to the task at hand may not be
immediate.
%

This paper addressed this problem by developing a general paradigm for
multi-class classification that sequentially restricts the set of
candidate classes to a small set, in a way that is driven by the data
observed.
We have described the method and provided some justifications for its
advantages, especially in NLP-like domains. Preliminary experiments
also show promise.

Several issues are still missing from this work.
In our experimental study the decomposition of the feature space was
done manually; it would be nice to develop methods to do this
automatically. Better understanding of methods for thresholding the
probability distributions that the classifiers output, as well as
principled ways to order them are also among the future directions of
this research.


\bibliographystyle{acl}
\bibliography{colt,learn,MyBib,nlp}

\end{document}